\documentclass[10pt,twocolumn,letterpaper]{article}

\usepackage{cvpr}
\usepackage{times}
\usepackage{epsfig}
\usepackage{graphicx}
\usepackage{amsmath}
\usepackage{amssymb}
\usepackage{multirow}
\usepackage{textcomp}
\usepackage{float}

\usepackage[pagebackref=true,breaklinks=true,letterpaper=true,colorlinks,bookmarks=false]{hyperref}

\cvprfinalcopy 

\def\cvprPaperID{2959} 

\ifcvprfinal\fi

\begin{document}

\title{Pyramid Stereo Matching Network}

\author{Jia-Ren Chang \,\,\,\,\,\,\,\,\,\,\,\,\,\,\,\,\,\,\,\,\,\,\,\,\, Yong-Sheng Chen\\
Department of Computer Science, National Chiao Tung University, Taiwan\\
{\tt\small \{followwar.cs00g, yschen\}@nctu.edu.tw}
}

\maketitle
\thispagestyle{empty}

\begin{abstract}
Recent work has shown that depth estimation from a stereo pair of images can be formulated as a supervised learning task to be resolved with convolutional neural networks (CNNs).
However, current architectures rely on patch-based Siamese networks, lacking the means to exploit context information for finding correspondence in ill-posed regions.
To tackle this problem, we propose PSMNet, a pyramid stereo matching network consisting of two main modules: spatial pyramid pooling and 3D CNN.
The spatial pyramid pooling module takes advantage of the capacity of global context information by aggregating context in different scales and locations to form a cost volume.
The 3D CNN learns to regularize cost volume using stacked multiple hourglass networks in conjunction with intermediate supervision.
The proposed approach was evaluated on several benchmark datasets.
Our method ranked first in the KITTI 2012 and 2015 leaderboards before March 18, 2018.
The codes of PSMNet are available at: \url{https://github.com/JiaRenChang/PSMNet}.

\end{abstract}

\section{Introduction}
Depth estimation from stereo images is essential to computer vision applications, including autonomous driving for vehicles, 3D model reconstruction, and object detection and recognition~\cite{chen20153d,zhang2015meshstereo}.
Given a pair of rectified stereo images, the goal of depth estimation is to compute the disparity $d$ for each pixel in the reference image.
Disparity refers to the horizontal displacement between a pair of corresponding pixels on the left and right images.
For the pixel $(x, y)$ in the left image, if its corresponding point is found at $(x - d, y)$ in the right image, then the depth of this pixel is calculated by $\frac{fB}{d}$, where $f$ is the camera\textquotesingle s focal length and $B$ is the distance between two camera centers.

The typical pipeline for stereo matching involves the finding of corresponding points based on matching cost and post-processing.
Recently, convolutional neural networks (CNNs) have been applied to learn how to match corresponding points in MC-CNN~\cite{zbontar2016stereo}.
Early approaches using CNNs treated the problem of correspondence estimation as similarity computation~\cite{Shaked_2017_CVPR, zbontar2016stereo}, where CNNs compute the similarity score for a pair of image patches to further determine whether they are matched.
Although CNN yields significant gains compared to conventional approaches in terms of both accuracy and speed, it is still difficult to find accurate corresponding points in inherently ill-posed regions such as occlusion areas, repeated patterns, textureless regions, and reflective surfaces.
Solely applying the intensity-consistency constraint between different viewpoints is generally insufficient for accurate correspondence estimation in such ill-posed regions, and is useless in textureless regions.
Therefore, regional support from global context information must be incorporated into stereo matching.

One major problem with current CNN-based stereo matching methods is how to effectively exploit context information.
Some studies attempt to incorporate semantic information to largely refine cost volumes or disparity maps~\cite{guney2015displets, Kendall_2017_ICCV, Shaked_2017_CVPR}.
The Displets~\cite{guney2015displets} method utilizes object information by modeling 3D vehicles to resolve ambiguities in stereo matching.
ResMatchNet~\cite{Shaked_2017_CVPR} learns to measure reflective confidence for the disparity maps to improve performance in ill-posed regions.
GC-Net~\cite{Kendall_2017_ICCV} employs the encoder-decoder architecture to merge multiscale features for cost volume regularization.

In this work, we propose a novel pyramid stereo matching network (PSMNet) to exploit global context information in stereo matching.
Spatial pyramid pooling (SPP)~\cite{he2014spatial,Zhao_2017_CVPR} and dilated convolution~\cite{chen2016deeplab, yu2015multi} are used to enlarge the receptive fields.
In this way, PSMNet extends pixel-level features to region-level features with different scales of receptive fields; the resultant combined global and local feature clues are used to form the cost volume for reliable disparity estimation.
Moreover, we design a stacked hourglass 3D CNN in conjunction with intermediate supervision to regularize the cost volume. The stacked hourglass 3D CNN repeatedly processes the cost volume in a top-down/bottom-up manner to further improve the utilization of global context information.

Our main contributions are listed below:
\begin{itemize}
\item We propose an end-to-end learning framework for stereo matching without any post-processing.
\item We introduce a pyramid pooling module for incorporating global context information into image features.
\item We present a stacked hourglass 3D CNN to extend the regional support of context information in cost volume.
\item We achieve state-of-the-art accuracy on the KITTI dataset.
\end{itemize}

\section{Related Work}
For depth estimation from stereo images, many methods for matching cost computation and cost volume optimization have been proposed in the literature.
According to~\cite{scharstein2002taxonomy}, a typical stereo matching algorithm consists of four steps: matching cost computation, cost aggregation, optimization, and disparity refinement.

Current state-of-the-art studies focus on how to accurately compute the matching cost using CNNs and how to apply semi-global matching (SGM)~\cite{hirschmuller2005accurate} to refine the disparity map.
Zbontar and LeCun~\cite{zbontar2016stereo} introduce a deep Siamese network to compute matching cost.
Using a pair of $9\times9$ image patches, the network is trained to learn to predict the similarity between image patches. 
Their method also exploits typical stereo matching procedures, including cost aggregation, SGM, and other disparity map refinements to improve matching results.
Further studies improve stereo depth estimation.
Luo~\etal \cite{luo2016efficient} propose a notably faster Siamese network in which the computation of matching costs is treated as a multi-label classification.
Shaked and Wolf~\cite{Shaked_2017_CVPR} propose a highway network for matching cost computation and a global disparity network for the prediction of disparity confidence scores, which facilitate the further refinement of disparity maps.

Some studies focus on the post-processing of the disparity map.
The Displets~\cite{guney2015displets} method is proposed based on the fact that objects generally exhibit regular structures, and are not arbitrarily shaped.
In the Displets~\cite{guney2015displets} method, 3D models of vehicles are used to resolve matching ambiguities in reflective and textureless regions.
Moreover, Gidaris and Komodakis~\cite{Gidaris_2017_CVPR} propose a network architecture which improves the labels by detecting incorrect labels, replacing incorrect labels with new ones, and refining the renewed labels (DRR).
Gidaris and Komodakis~\cite{Gidaris_2017_CVPR} use the DRR network on disparity maps and achieve good performance without other post-processing.
The SGM-Net~\cite{Seki_2017_CVPR} learns to predict SGM penalties instead of manually-tuned penalties for regularization.

Recently, end-to-end networks have been developed to predict whole disparity maps without post-processing.
Mayer~\etal~\cite{Mayer_2016_CVPR} present end-to-end networks for the estimation of disparity (DispNet) and optical flow (FlowNet).
They also offer a large synthetic dataset, Scene Flow, for network training.
Pang~\etal~\cite{Pang_2017_ICCV_Workshops} extend DispNet~\cite{Mayer_2016_CVPR} and introduce a two-stage network called cascade residual learning (CRL).
The first and second stages calculate the disparity map and its multi-scale residuals, respectively.
Then the outputs of both stages are summed to form the final disparity map.
Also, Kendall~\etal~\cite{Kendall_2017_ICCV} introduce GC-Net, an end-to-end network for cost volume regularization using 3D convolutions.
The above-mentioned end-to-end approaches exploit multiscale features for disparity estimation.
Both DispNet~\cite{Mayer_2016_CVPR} and CRL~\cite{Pang_2017_ICCV_Workshops} reuse hierarchical information, concatenating features from lower layers with those from higher layers.
CRL~\cite{Pang_2017_ICCV_Workshops} also uses hierarchical supervision to calculate disparity in multiple resolutions.
GC-Net~\cite{Kendall_2017_ICCV} applies the encoder-decoder architecture to regularize the cost volume.
The main idea of these methods is to incorporate context information to reduce mismatch in ambiguous regions and thus improve depth estimation.

In the field of semantic segmentation, aggregating context information is also essential for labeling object classes.
There are two main approaches to exploiting global context information: the encoder-decoder architecture and pyramid pooling.
The main idea of the encoder-decoder architecture is to integrate top-down and bottom-up information via skip connections.
The fully convolutional network (FCN)~\cite{long2015fully} was first proposed to aggregate coarse-to-fine predictions to improve segmentation results.
U-Net~\cite{ronneberger2015u}, instead of aggregating coarse-to-fine predictions, aggregates coarse-to-fine features and achieves good segmentation results for biomedical images.
Further studies including SharpMask~\cite{pinheiro2016learning}, RefineNet~\cite{Lin_2017_CVPR}, and the label refinement network~\cite{islam2017label} follow this core idea and propose more complex architectures for the merging of multiscale features.
Moreover, stacked multiple encoder-decoder networks such as~\cite{fu2017stacked} and~\cite{newell2016stacked} were introduced to improve feature fusion.
In~\cite{newell2016stacked}, the encoder-decoder architecture is termed the \textit{hourglass} architecture.

Pyramid pooling was proposed based on the fact that the empirical receptive field is much smaller than the theoretical receptive field in deep networks~\cite{liu2015parsenet}.
ParseNet~\cite{liu2015parsenet} demonstrates that global pooling with FCN enlarges the empirical receptive field to extract information at the whole-image level and thus improves semantic segmentation results.
DeepLab v2~\cite{chen2016deeplab} proposes atrous spatial pyramid pooling (ASPP) for multiscale feature embedding, containing parallel dilated convolutions with different dilated rates.
PSPNet~\cite{Zhao_2017_CVPR} presents a pyramid pooling module to collect the effective multiscale contextual prior.
Inspired by PSPNet~\cite{Zhao_2017_CVPR}, DeepLab v3~\cite{chen2017rethinking} proposes a new ASPP module augmented with global pooling.

Similar ideas of spatial pyramids have been used in context of optical flow. 
SPyNet~\cite{ranjan2017optical} introduces image pyramids to estimate optical flow in a coarse-to-fine approach.
PWCNet~\cite{sun2017pwc} improves optical flow estimation by using feature pyramids. 

In this work on stereo matching, we embrace the experience of semantic segmentation studies and exploit global context information at the whole-image level.
As described below, we propose multiscale context aggregation via a pyramid stereo matching network for depth estimation.

\begin{figure*}[]
\begin{center}
	\includegraphics*[width=\textwidth]{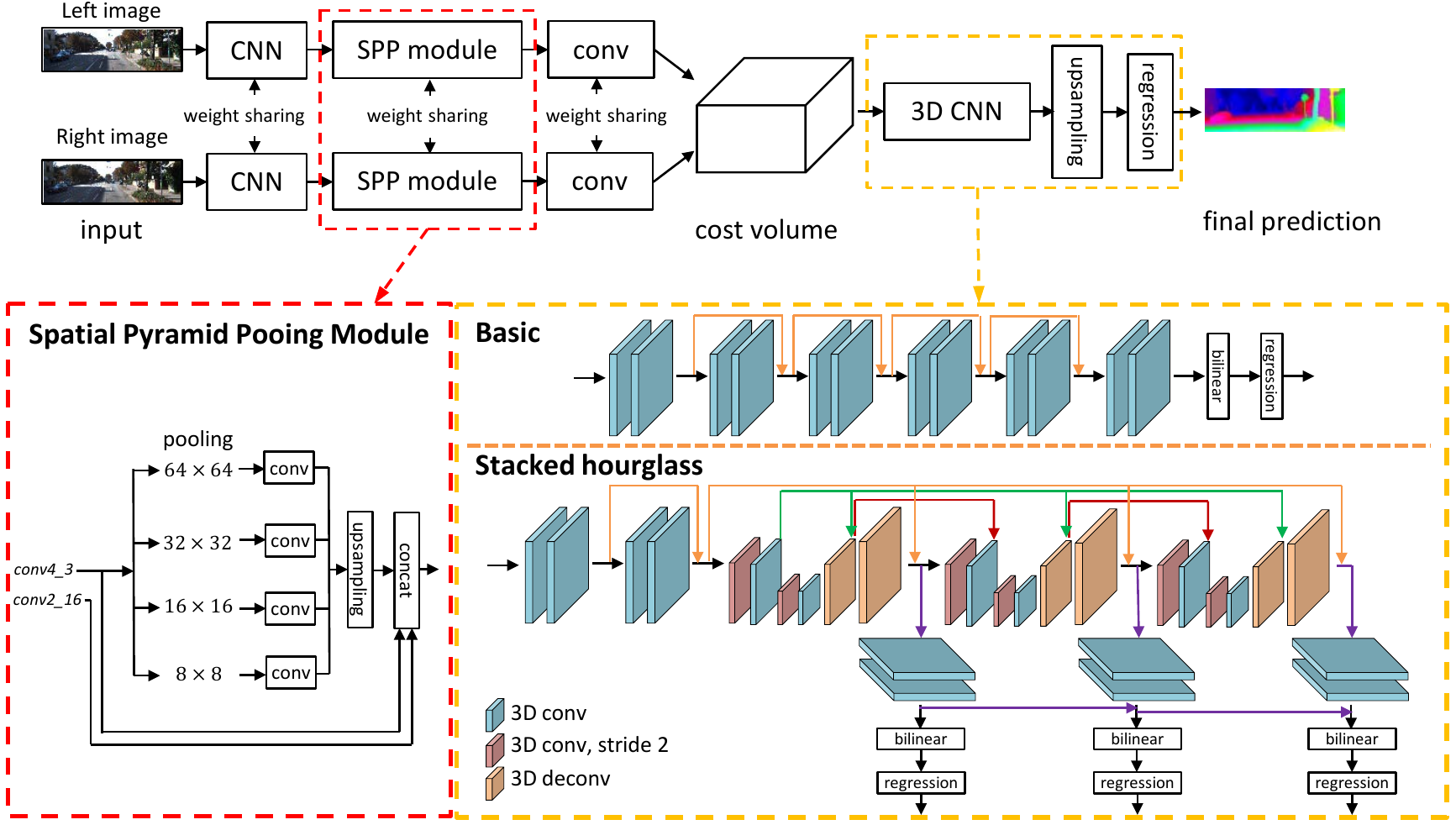}
\end{center}
   \caption{Architecture overview of proposed PSMNet. The left and right input stereo images are fed to two weight-sharing pipelines consisting of a CNN for feature maps calculation, an SPP module for feature harvesting by concatenating representations from sub-regions with different sizes, and a convolution layer for feature fusion. The left and right image features are then used to form a 4D cost volume, which is fed into a 3D CNN for cost volume regularization and disparity regression.}
\label{Architecture}
\end{figure*}

\section{Pyramid Stereo Matching Network}
We present PSMNet, which consists of an SPP~\cite{he2014spatial, Zhao_2017_CVPR} module for effective incorporation of global context and a stacked hourglass module for cost volume regularization.
The architecture of PSMNet is illustrated in Figure~\ref{Architecture}.

\subsection{Network Architecture}
The parameters of the proposed PSMNet are detailed in Table~\ref{Tab1Parameters}.
In contrast to the application of large filters ($7\times7$) for the first convolution layer in other studies~\cite{he2016deep}, three small convolution filters ($3\times3$) are cascaded to construct a deeper network with the same receptive field.
The conv1\_x, conv2\_x, conv3\_x, and conv4\_x are the basic residual blocks~\cite{he2016deep} for learning the unary feature extraction.
For conv3\_x and conv4\_x, dilated convolution is applied to further enlarge the receptive field.
The output feature map size is  $\frac{1}{4}\times\frac{1}{4}$ of the input image size, as shown in Table~\ref{Tab1Parameters}.
The SPP module, as shown in Figure~\ref{Architecture}, is then applied to gather context information.
We concatenate the left and right feature maps into a cost volume, which is fed into a 3D CNN for regularization.
Finally, regression is applied to calculate the output disparity map.
The SPP module, cost volume, 3D CNN, and disparity regression are described in later sections.

\begin{table}
\centering
   \caption{Parameters of the proposed PSMNet architecture. Construction of residual blocks are designated in brackets with the number of stacked blocks. Downsampling is performed by conv0\_1 and conv2\_1 with stride of 2. The usage of batch normalization and ReLU follows ResNet~\cite{he2016deep}, with exception that PSMNet does not apply ReLU after summation. $H$ and $W$ denote the height and width of the input image, respectively, and $D$ denotes the maximum disparity.}
\begin{center}
	\includegraphics[width=2.2in]{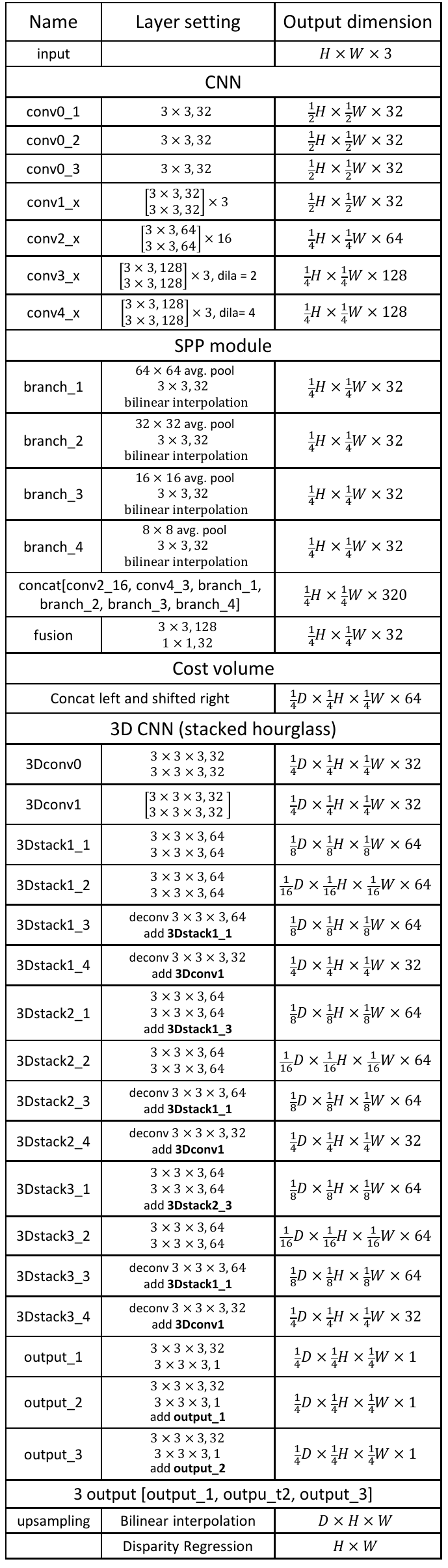}
\end{center}
\label{Tab1Parameters}
\end{table}

\subsection{Spatial Pyramid Pooling Module}
It is difficult to determine the context relationship solely from pixel intensities.
Therefore, image features rich with object context information can benefit correspondence estimation, particularly for ill-posed regions.
In this work, the relationship between an object (for example, a car) and its sub-regions (windows, tires, hoods, etc.) is learned by the SPP module to incorporate hierarchical context information.

In~\cite{he2014spatial}, SPP was designed to remove the fixed-size constraint of CNN.
Feature maps at different levels generated by SPP are flattened and fed into the fully connected layer for classification, after which SPP is applied to semantic segmentation problems.
ParseNet~\cite{liu2015parsenet} applies global average pooling to incorporate global context information.
PSPNet~\cite{Zhao_2017_CVPR} extends ParseNet~\cite{liu2015parsenet} to a hierarchical global prior, containing information with different scales and sub-regions.
In~\cite{Zhao_2017_CVPR}, the SPP module uses adaptive average pooling to compress features into four scales and is followed by a $1\times1$ convolution to reduce feature dimension, after which the low-dimensional feature maps are upsampled to the same size of the original feature map via bilinear interpolation.
The different levels of feature maps are concatenated as the final SPP feature maps.

In the current work, we design four fixed-size average pooling blocks for SPP: $64\times64$, $32\times32$, $16\times16$, and $8\times8$, as shown in Figure~\ref{Architecture} and Table~\ref{Tab1Parameters}.
Further operations, including $1\times1$ convolution and upsampling, are the same as in~\cite{Zhao_2017_CVPR}.
In an ablation study, we performed extensive experiments to show the effect of feature maps at different levels, as described in Section 4.2.

\subsection{Cost Volume}
Rather than using a distance metric, the MC-CNN~\cite{zbontar2016stereo} and GC-Net~\cite{Kendall_2017_ICCV} approaches concatenate the left and right features to learn matching cost estimation using deep network.
Following~\cite{Kendall_2017_ICCV}, we adopt SPP features to form a cost volume by concatenating left feature maps with their corresponding right feature maps across each disparity level, resulting in a 4D volume (height$\times$width$\times$disparity$\times$feature size).

\subsection{3D CNN}
The SPP module facilitates stereo matching by involving different levels of features.
To aggregate the feature information along the disparity dimension as well as spatial dimensions, we propose two kinds of 3D CNN architectures for cost volume regularization: the basic and stacked hourglass architectures.
In the basic architecture, as shown in Figure\ref{Architecture}, the network is simply built using residual blocks.
The basic architecture contains twelve  $3\times3\times3$ convolutional layers.
Then we upsample the cost volume back to size $H\times W\times D$ via bilinear interpolation.
Finally, we apply regression to calculate the disparity map with size $H\times W$, which is introduced in Section 3.5. 

In order to learn more context information, we use a stacked hourglass (encoder-decoder) architecture, consisting of repeated top-down/bottom-up processing in conjunction with intermediate supervision, as shown in Figure~\ref{Architecture}.
The stacked hourglass architecture has three main hourglass networks, each of which generates a disparity map.
That is, the stacked hourglass architecture has three outputs and losses (Loss\_1, Loss\_2, and Loss\_3).
The loss function is described in Section 3.6.
During the training phase, the total loss is calculated as the weighted summation of the three losses.
During the testing phase, the final disparity map is the last of three outputs.
In our ablation study, the basic architecture was used to evaluate the performance of the SPP module, because the basic architecture does not learn extra context information through the encoding/decoding process as in~\cite{Kendall_2017_ICCV}.

\subsection{Disparity Regression}
We use disparity regression as proposed in~\cite{Kendall_2017_ICCV} to estimate the continuous disparity map.
The probability of each disparity $d$ is calculated from the predicted cost $c_d$ via the softmax operation $\sigma(\cdot)$.
The predicted disparity $\hat{d}$ is calculated as the sum of each disparity $d$ weighted by its probability, as
\begin{equation}
\hat{d} = \sum_{d=0}^{D_{max}}{d\times\sigma(-c_d)}.
\end{equation}

As reported in~\cite{Kendall_2017_ICCV}, the above disparity regression is more robust than classification-based stereo matching methods.
Note that the above equation is similar to that introduced in~\cite{bahdanau2014neural}, in which it is referred to as a soft attention mechanism. 

\subsection{Loss}
Because of the disparity regression, we adopt the smooth $L_1$ loss function to train the proposed PSMNet.
Smooth $L_1$ loss is widely used in bounding box regression for object detection because of its robustness and low sensitivity to outliers~\cite{girshick2015fast}, as compared to $L_2$ loss. The loss function of PSMNet is defined as
\begin{equation}
L(d, \hat{d}) = {\frac{1}{N}}\sum_{i=1}^{N}{smooth_{L_1}(d_i-\hat{d_i})},
\end{equation}
in which

\[
smooth_{L_1}(x) = \left\{\begin{array}{ll}
                 0.5x^2, & \mbox{if $|x|<1$} \\  
                 |x|-0.5, & \mbox{otherwise}  
                \end{array} \right.,
\]
where $N$ is the number of labeled pixels, $d$ is the ground-truth disparity, and $\hat{d}$ is the predicted disparity.

\section{Experiments}
We evaluated our method on three stereo datasets: Scene Flow, KITTI 2012, and KITTI 2015.
We also performed ablation studies using KITTI 2015 with our architecture setting to evaluate the influence on performance made by dilated convolution, different sizes of pyramid pooling, and the stacked hourglass 3D CNN.
The experimental settings and network implementation are presented in Section 4.1, followed by the evaluation results on each of the three stereo datasets used in this study. 

\begin{table*}
\centering
   \caption{Evaluation of PSMNet with different settings. We computed the percentage of three-pixel-error on the KITTI 2015 validation set, and end-point-error on the Scene Flow test set. * denote that we use half the dilated rate of dilated convolution.}
\begin{center}
\begin{tabular}{|c|cccc|c|c|c|}
\hline
\multicolumn{6}{|c|}{Network setting} & KITTI 2015 & Scene Flow  \\
\hline
\multirow{2}{*}{dilated conv} & \multicolumn{4}{c}{pyramid pooling size} \vline & \multirow{2}{*}{stacked hourglass} & \multirow{2}{*}{Val Err (\%)} & \multirow{2}{*}{End Point Err} \\ 
\cline{2-5}
&$64\times64$&$32\times32$&$16\times16$&$8\times8$& & & \\ \hline

&&&&&&2.43&1.43\\ \hline
${\surd}$&&&&&&2.16&1.56\\ \hline
&${\surd}$&${\surd}$&${\surd}$&${\surd}$&&2.47&1.40\\ \hline
${\surd}$&${\surd}$&&&&&2.17&1.30\\ \hline
${\surd}$&${\surd}$&${\surd}$&${\surd}$&${\surd}$&&2.09&1.28\\ \hline
${\surd}$&${\surd}$&${\surd}$&${\surd}$&${\surd}$&${\surd}$&1.98&\textbf{1.09}\\ \hline
${\surd}$*&${\surd}$&${\surd}$&${\surd}$&${\surd}$&${\surd}$&\textbf{1.83}&1.12\\ \hline
\end{tabular}
\end{center}
\label{ablationsetting}
\end{table*}

\subsection{Experiment Details}
We evaluated our method on three stereo datasets:
\begin{enumerate}
\item Scene Flow: a large scale synthetic dataset containing 35454 training and 4370 testing images with $H=540$ and $W=960$.
This dataset provides dense and elaborate disparity maps as ground truth.
Some pixels have large disparities and are excluded in the loss computation if the disparity is larger than the limits set in our experiment.
\item KITTI 2015: a real-world dataset with street views from a driving car.
It contains 200 training stereo image pairs with sparse ground-truth disparities obtained using LiDAR and another 200 testing image pairs without ground-truth disparities.
Image size is $H=376$ and $W=1240$.
We further divided the whole training data into a training set (80\%) and a validation set (20\%).
\item KITTI 2012: a real-world dataset with street views from a driving car.
It contains 194 training stereo image pairs with sparse ground-truth disparities obtained using LiDAR and 195 testing image pairs without ground-truth disparities.
Image size is $H=376$ and $W=1240$.
We further divided the whole training data into a training set (160 image pairs) and a validation set (34 image pairs).
Color images of KITTI 2012 were adopted in this work.
\end{enumerate}

The full architecture of the proposed PSMNet is shown in Table~\ref{Tab1Parameters}, including the number of convolutional filters.
The usage of batch normalization and ReLU is the same as in ResNet~\cite{he2016deep}, with exception that PSMNet does not apply ReLU after summation.

The PSMNet architecture was implemented using PyTorch.
All models were end-to-end trained with Adam ($\beta_1=0.9,\beta_2=0.999$).
We performed color normalization on the entire dataset for data preprocessing.
During training, images were randomly cropped to size $H=256$ and $W=512$.
The maximum disparity ($D$) was set to 192.
We trained our models from scratch using the Scene Flow dataset with a constant learning rate of 0.001 for 10 epochs.
For Scene Flow, the trained model was directly used for testing.
For KITTI, we used the model trained with Scene Flow data after fine-tuning on the KITTI training set for 300 epochs.
The learning rate of this fine-tuning began at 0.001 for the first 200 epochs and 0.0001 for the remaining 100 epochs.
The batch size was set to 12 for the training on four nNvidia Titan-Xp GPUs (each of 3).
The training process took about 13 hours for Scene Flow dataset and 5 hours for KITTI datasets.
Moreover, we prolonged the training process to 1000 epochs to obtain the final model and the test results for KITTI submission.

\subsection{KITTI 2015}

\paragraph{Ablation study for PSMNet}
We conducted experiments with several settings to evaluate PSMNet, including the usage of dilated convolution, pooling at different levels, and 3D CNN architectures.
The default 3D CNN design was the basic architecture.
As listed in Table~\ref{ablationsetting}, dilated convolution works better when used in conjunction with the SPP module.
For pyramid pooling, pooling with more levels works better.
The stacked hourglass 3D CNN significantly outperformed the basic 3D CNN when combined with dilated convolution and the SPP module.
The best setting of PSMNet yielded a 1.83\% error rate on the KITTI 2015 validation set.

\paragraph{Ablation study for Loss Weight}
The stacked hourglass 3D CNN has three outputs for training and can facilitate the learning process.
As shown in Table~\ref{ablationweight}, we conducted experiments with various combinations of loss weights between 0 and 1.
For the baseline, we treated the three losses equally and set all to 1.
The results showed that the weight settings of 0.5 for Loss\_1, 0.7 for Loss\_2, and 1.0 for Loss\_3 yielded the best performance, which was a 1.98\% error rate on the KITTI 2015 validation set.

\begin{table}
\centering
   \caption{ Influence of weight values for Loss\_1, Loss\_2, and Loss\_3 on validation errors. We empirically found that 0.5/0.7/1.0 yielded the best performance.}
\begin{center}
\begin{tabular}{|ccc|c|}
\hline
\multicolumn{3}{|c|}{Loss weight} & KITTI 2015  \\ 
\cline{1-3}
Loss\_1 & Loss\_2 & Loss\_3 &val error(\%) \\ \hline
0.0 & 0.0 & 1.0 & 2.49 \\ \hline
0.1 & 0.3 & 1.0 & 2.07 \\ \hline
0.3 & 0.5 & 1.0 & 2.05 \\ \hline
0.5 & 0.7 & 1.0 & \textbf{1.98} \\ \hline
0.7 & 0.9 & 1.0 & 2.05 \\ \hline
1.0 & 1.0 & 1.0 & 2.01 \\ \hline

\end{tabular}
\end{center}
\label{ablationweight}
\end{table}

\begin{table*}[t]
\centering
   \caption{The KITTI 2015 leaderboard presented on March 18, 2018. The results show the percentage of pixels with errors of more than three pixels or 5\% of disparity error from all test images. Only published methods are listed for comparison.}
\begin{center}
\begin{tabular}{|c|c|c|c|c|c|c|c|c|}
\hline
\multirow{2}{*}{Rank} & \multirow{2}{*}{Method} & \multicolumn{3}{c}{All (\%)} \vline & \multicolumn{3}{c}{Noc (\%)} \vline & \multirow{2}{*}{Runtime (s)}  \\
\cline{3-8}
&  &D1-bg& D1-fg&D1-all & D1-bg&D1-fg&D1-all&  \\ \hline

1 & PSMNet (ours) &\textbf{1.86}& 4.62 & \textbf{2.32} & \textbf{1.71} & 4.31 & \textbf{2.14} & 0.41 \\ \hline
3 & iResNet-i2e2~\cite{liang2017learning} & 2.14 & 3.45 & 2.36 & 1.94 & 3.20 & 2.15 & 0.22 \\ \hline
6 & iResNet~\cite{liang2017learning} & 2.35 & \textbf{3.23} & 2.50 & 2.15 & \textbf{2.55} & 2.22 & \textbf{0.12} \\ \hline
8 & CRL~\cite{Pang_2017_ICCV_Workshops} & 2.48 & 3.59 & 2.67 & 2.32 & 3.12 & 2.45 & 0.47 \\ \hline
11 & GC-Net~\cite{Kendall_2017_ICCV} & 2.21 & 6.16 & 2.87 & 2.02 & 5.58 & 2.61 & 0.90 \\ \hline
\end{tabular}
\end{center}
\label{2015leaderboard}
\end{table*}

\paragraph{Results on Leaderboard}
Using the best model trained in our experiments, we calculated the disparity maps for the 200 testing images in the KITTI 2015 dataset and submitted the results to the KITTI evaluation server for the performance evaluation.
According to the online leaderboard, as shown in Table~\ref{2015leaderboard}, the overall three-pixel-error for the proposed PSMNet was \textbf{2.32}\%, which surpassed prior studies by a noteworthy margin.
In this table, ``All'' means that all pixels were considered in error estimation, whereas ``Noc'' means that only the pixels in non-occluded regions were taken into account.
The three columns ``D1-bg'', ``D1-fg'' and ``D1-all'' mean that the pixels in the background, foreground, and all areas, respectively, were considered in the estimation of errors.

\begin{figure*}
\begin{center}
	\includegraphics*[width=\textwidth]{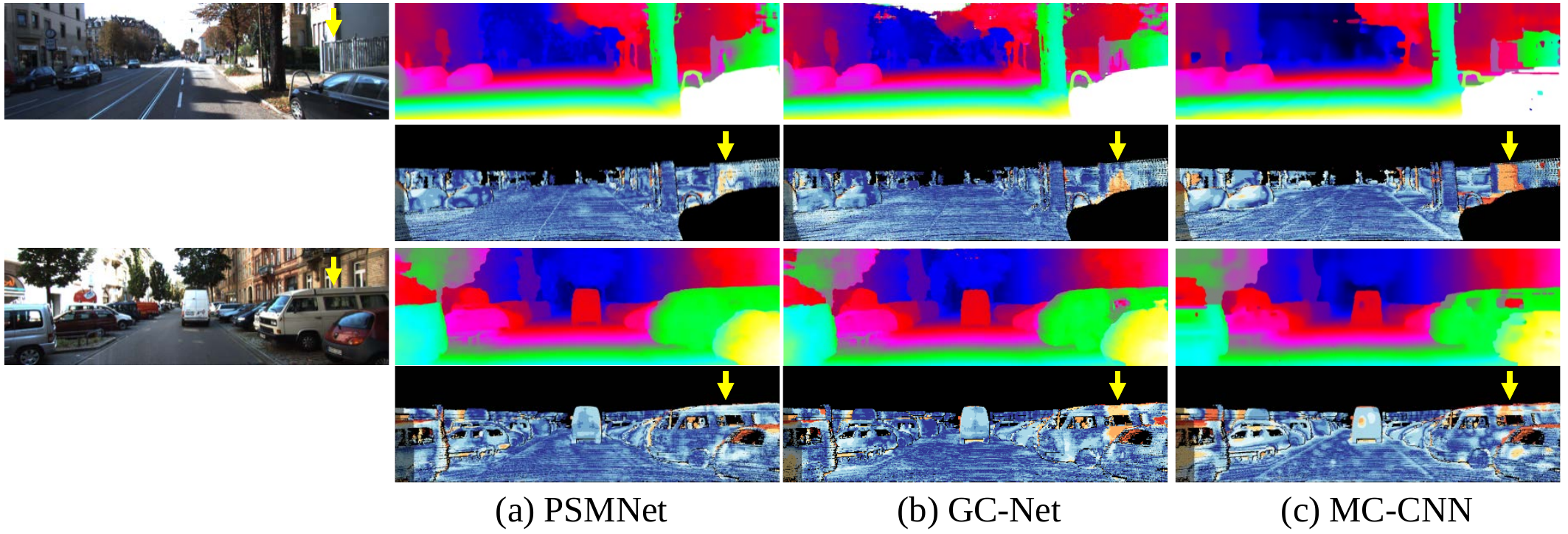}
\end{center}
   \caption{Results of disparity estimation for KITTI 2015 test images. The left panel shows the left input image of stereo image pair. For each input image, the disparity maps obtained by (a) PSMNet, (b) GC-Net~\cite{Kendall_2017_ICCV}, and (c) MC-CNN~\cite{zbontar2016stereo} are illustrated together above their error maps.}
\label{2015results}
\end{figure*}

\paragraph{Qualitative evaluation}
Figure~\ref{2015results} illustrates some examples of the disparity maps estimated by the proposed PSMNet, GC-Net~\cite{Kendall_2017_ICCV}, and MC-CNN~\cite{zbontar2016stereo} together with the corresponding error maps.
These results were reported by the KITTI evaluation server.
PSMNet yields more robust results, particularly in ill-posed regions, as indicated by the yellow arrows in Figure~\ref{2015results}.
Among these three methods, PSMNet more correctly predicts the disparities for the fence region, indicated by the yellow arrows in the middle row of Figure~\ref{2015results}.

\subsection{Scene Flow}
\begin{figure*}
\begin{center}
	\includegraphics*[width=0.96\textwidth]{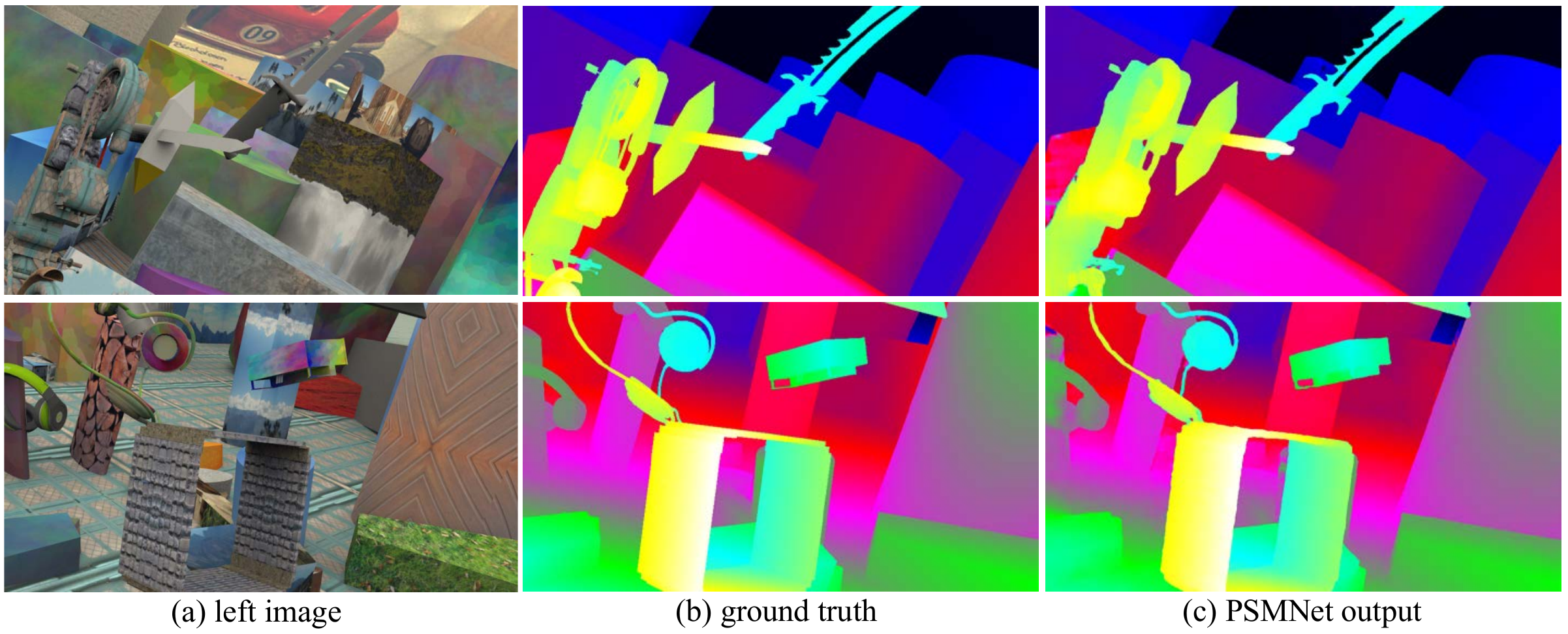}
\end{center}
   \caption{Performance evaluation using Scene Flow test data. (a) left image of stereo image pair, (b) ground truth disparity, and (c) disparity map estimated using PSMNet.}
\label{sceneflow}
\end{figure*}

We also compared the performance of PSMNet with other state-of-the-art methods, including CRL~\cite{Pang_2017_ICCV_Workshops}, DispNetC~\cite{Mayer_2016_CVPR}, GC-Net~\cite{Kendall_2017_ICCV}, using the Scene Flow test set.
As shown in Table~\ref{scenecomp}, PSMNet outperformed other methods in terms of accuracy.
Three testing examples are illustrated in Figure~\ref{sceneflow} to demonstrate that PSMNet obtains accurate disparity maps for delicate and intricately overlapped objects. 

\begin{table}
\centering
   \caption{Performance comparison with Scene Flow test set. EPE: End-point-error.}
\begin{center}
\resizebox{\linewidth}{!}{\begin{tabular}{|c|c|c|c|c|c|}
\hline
& PSMNet & CRL~\cite{Pang_2017_ICCV_Workshops} & DispNetC~\cite{Mayer_2016_CVPR} & GC-Net~\cite{Kendall_2017_ICCV}  \\ \hline
EPE & \textbf{1.09} & 1.32 &1.68 &2.51 \\ \hline
\end{tabular}}
\end{center}
\label{scenecomp}
\end{table}

\subsection{KITTI 2012}
Using the best model trained in our experiments, we calculated the disparity maps for the 195 testing images in the KITTI 2012 dataset and submitted the results to the KITTI evaluation server for the performance evaluation.
According to the online leaderboard, as shown in Table~\ref{2012comp}, the overall three-pixel-error for the proposed PSMNet was \textbf{1.89}\%, which surpassed prior studies by a noteworthy margin.

\begin{table*}
\centering
   \caption{The leaderboard of KITTI 2012 presented on March 18, 2018. PSMNet achieves the best results under all evaluation criteria, except runtime. Only published methods are listed for comparison.}
\begin{center}
\begin{tabular}{|c|c|c|c|c|c|c|c|c|c|c|}
\hline
\multirow{2}{*}{Rank} & \multirow{2}{*}{Method} & \multicolumn{2}{c}{$>$2 px} \vline & \multicolumn{2}{c}{$>$3 px} \vline & \multicolumn{2}{c}{$>$5 px} \vline & \multicolumn{2}{c}{Mean Error} \vline & \multirow{2}{*}{Runtime (s)}  \\
\cline{3-10}
&  &Noc& All & Noc & All &Noc&All&Noc&All&  \\ \hline

1 & PSMNet (ours) &\textbf{2.44}&\textbf{3.01}&\textbf{1.49}&\textbf{1.89}&\textbf{0.90}&\textbf{1.15}&\textbf{0.5}&\textbf{0.6}& 0.41 \\ \hline
2 & iResNet-i2~\cite{liang2017learning} &2.69&3.34&1.71&2.16&1.06&1.32&0.5&0.6&\textbf{0.12} \\ \hline
4 & GC-Net~\cite{Kendall_2017_ICCV} &2.71&3.46&1.77&2.30&1.12&1.46&0.6&0.7&0.9 \\ \hline
11 & L-ResMatch~\cite{Shaked_2017_CVPR} &3.64&5.06&2.27&3.40&1.50&2.26&0.7&1.0&48 \\ \hline
14 & SGM-Net~\cite{Seki_2017_CVPR} &3.60&5.15&2.29&3.50&1.60&2.36&0.7&0.9&67 \\ \hline
\end{tabular}
\end{center}
\label{2012comp}
\end{table*}

\begin{figure*}[!btp]
\begin{center}
	\includegraphics*[width=0.96\textwidth]{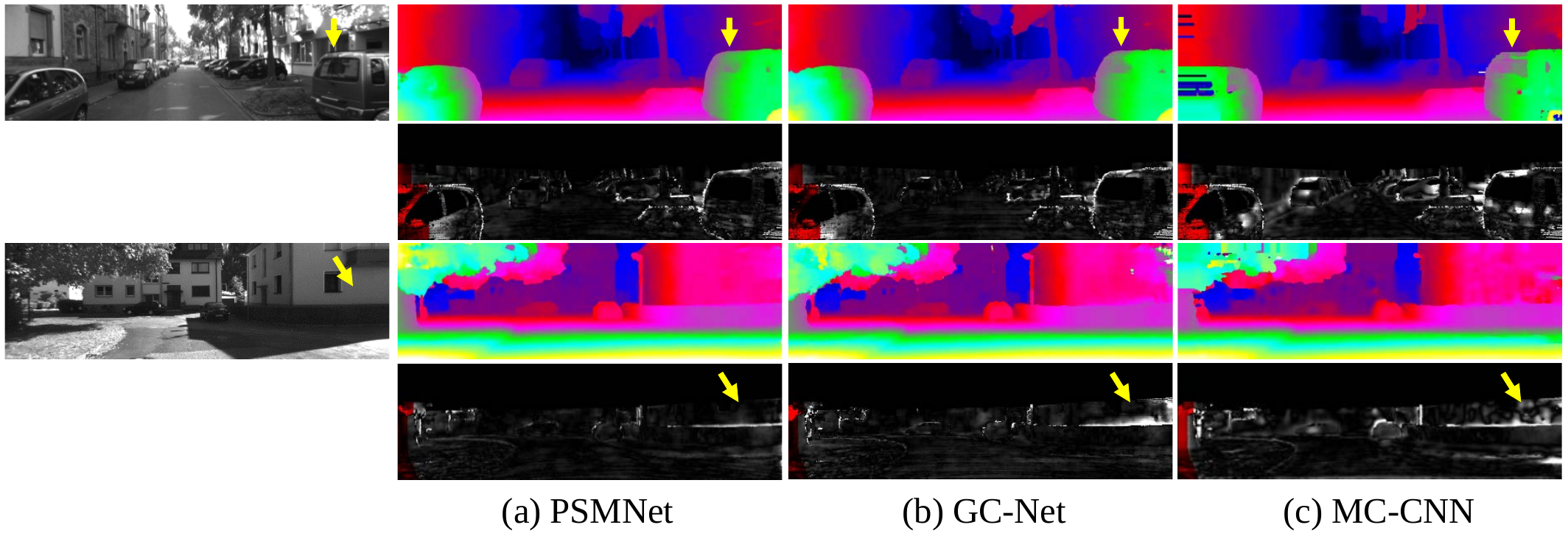}
\end{center}
   \caption{Results of disparity estimation for KITTI 2012 test images. The left panel shows the left input image of the stereo image pair. For each input image, the disparity obtained by (a) PSMNet, (b) GC-Net~\cite{Kendall_2017_ICCV}, and (c) MC-CNN~\cite{zbontar2016stereo}, is illustrated above its error map.}
\label{2012result}
\end{figure*}

\paragraph{Qualitative evaluation}
Figure~\ref{2012result} illustrates some examples of the disparity maps estimated by the proposed PSMNet, GC-Net~\cite{Kendall_2017_ICCV}, and MC-CNN~\cite{zbontar2016stereo} together with the corresponding error maps.
These results were reported by the KITTI evaluation server.
PSMNet obtains more robust results, particularly in regions of car windows and walls, as indicated by the yellow arrows in Figure~\ref{2012result}.


\section{Conclusions}
Recent studies using CNNs for stereo matching have achieved prominent performance.
Nevertheless, it remains intractable to estimate disparity for inherently ill-posed regions.
In this work, we propose PSMNet, a novel end-to-end CNN architecture for stereo vision which consists of two main modules to exploit context information: the SPP module and the 3D CNN.
The SPP module incorporates different levels of feature maps to form a cost volume.
The 3D CNN further learns to regularize the cost volume via repeated top-down/bottom-up processes.
In our experiments, PSMNet outperforms other state-of-the-art methods.
PSMNet ranked first in both KITTI 2012 and 2015 leaderboards before March 18, 2018.
The estimated disparity maps clearly demonstrate that PSMNet significantly reduces errors in ill-posed regions.

\section*{Acknowledgement}
This work was supported in part by the Taiwan Ministry of Science and Technology (Grants MOST-106-2221-E-009-164-MY2 and MOST-107-2634-F-009-009).

{\small
\bibliographystyle{ieee}
\bibliography{egbib}
}

\end{document}